\documentclass[letterpaper, 10 pt, conference]{ieeeconf}  % Comment this line out if you need a4paper
\IEEEoverridecommandlockouts                              % This command is only needed if 
\usepackage[utf8]{inputenc}
\usepackage{graphicx}
\usepackage{algorithm}
\usepackage{algorithmic}
\usepackage{amsmath,amssymb,amsfonts}
\usepackage{commath}
\usepackage{xcolor}
\usepackage{hyperref}
\usepackage[ruled,vlined,linesnumbered, algo2e]{algorithm2e} 
\usepackage{multirow}
\usepackage{subcaption}
\usepackage{cite}
\hypersetup{
	colorlinks=true,
	linkcolor=blue,
	citecolor=red,
}

\newcommand{\scalemath}[2]{\scalebox{#1}{\mbox{\ensuremath{\displaystyle #2}}}}

\title{{\small \vspace{-2cm} {Accepted in IEEE International Conference on Robotics and Automation (ICRA 2021)}}\\ \vspace{1.5cm}\LARGE \bf
	Self-Imitation Learning by Planning
}

\author{Sha Luo, Hamidreza Kasaei, Lambert Schomaker% <-this % stops a space
	\thanks{All authors are with the Department of Artificial Intelligence, University of Groningen, the Netherlands. Sha Luo is funded by the China Scholarship Council. {\tt\small s.luo@rug.nl}}%
}
\def\BibTeX{{\rm B\kern-.05em{\sc i\kern-.025em b}\kern-.08em
		T\kern-.1667em\lower.7ex\hbox{E}\kern-.125emX}}
\begin{document}

\maketitle
\thispagestyle{empty}
\pagestyle{empty}
%%%%%%%%%%%%%%%%%%%%%%%%%%%%%%%%%%%%%%%%%%%%%%%%%%%%%%%%%%%%%%%%%%%%%%%%%%%%%%%%
\begin{abstract}

Imitation learning~(IL) enables robots to acquire skills quickly by transferring expert knowledge, which is widely adopted in reinforcement learning~(RL) to initialize exploration. However, in long-horizon motion planning tasks, a challenging problem in deploying IL and RL methods is how to generate and collect massive, broadly distributed data such that these methods can generalize effectively. In this work, we solve this problem using our proposed approach called {\em self-imitation learning by planning (SILP)}, where demonstration data are collected automatically by planning on the visited states from the current policy. SILP is inspired by the observation that successfully visited states in the early reinforcement learning stage are collision-free nodes in the graph-search based motion planner, so we can plan and relabel robot's own trials as demonstrations for policy learning. Due to these self-generated demonstrations, we relieve the human operator from the laborious data preparation process required by IL and RL methods in solving complex motion planning tasks. The evaluation results show that our SILP method achieves higher success rates and enhances sample efficiency compared to selected baselines, and the policy learned in simulation performs well in a real-world placement task with changing goals and obstacles. 
\end{abstract}

%%%%%%%%%%%%%%%%%%%%%%%%%%%%%%%%%%%%%%%%%%%%%%%%%%%%%%%%%%%%%%%%%%%%%%%%%%%%%%%%
\section{INTRODUCTION}

As a fundamental component, motion planning is deployed in many robotic platforms \cite{abbeel2008apprenticeship,chiang2019learning,rosenbaum2001posture,8794317,4543471}. Recently, there is an increasing interest in using imitation learning~(IL) and reinforcement learning~(RL) in motion planning~\cite{DBLP:conf/rss/JurgensonT19,qureshi2019motion, ravichandar2020recent}, driven by the demand of robots that can generalize well and react fast in dynamic environments. The primary difficulty of training neural-based motion planners in high-dimensional manipulation tasks lies in the data collection, as neural networks require massive diverse data for both IL and RL to generalize well. However, for IL, there is a shortage of data distribution around the boundary of obstacles when the goal is to provide obstacles-avoiding behaviors~\cite{DBLP:conf/rss/JurgensonT19}. In addition, a robot's performance is limited by the quality of supervision when there is no additional information to help improve its performance \cite{ross2010efficient}. For RL, exploring areas around the obstacles is unsafe in the real-world and it takes a long time to collect sufficient experience for training the policy to perform well in a dynamic environment with changing goals and obstacles. The combination of IL and RL can boost the performance in IL attributed to the improved exploration, and speed up the convergence in RL by transferring expert knowledge. However, to bootstrap the approach, the heavy data preparation process still needs to be realized, e.g., by using a simulated planner~\cite{chitta2012moveit}, demonstrating by human~\cite{rajeswaran2017learning,sauser2012iterative,kober2009policy} or collecting data in parallel with multiple robots~\cite{kalashnikov2018qt}. 

In this study, we propose a method to relieve human from the laborious data collection process. Reinforcement learning is a process intertwined with exploration and exploitation, and most trials in the early stage are not being utilized efficiently when they do not lead to success. Still, such experiences could provide beneficial biases towards obstacle-avoiding behaviors. Therefore, we regard these explored states as {\em candidate collision-free nodes} for the graph-based motion planning algorithms. By planning on these nodes, we generate demonstrations to guide the RL continuously. As we plan demonstrations online automatically by utilizing the self-generated states from the current policy, we name our approach as {\em self-imitation learning by planning (SILP)}. 

Our main contribution is the SILP method for online demonstrations collection to assist reinforcement learning in motion planning tasks while not introducing much extra computational burden on the training. The performance is improved automatically without human intervention. Aside from this, we mitigate the data discrepancy between demonstrations and interaction experience by planning on the visited states. The empirical results prove that our method achieves a higher success rate and better sample efficiency compared to other baselines. Furthermore, the policy learned in simulation can transfer well to real-world robot tasks in a dynamic environment. 
\begin{figure}[t]
	% \vspace{2mm}
	\centering
	\begin{subfigure}{0.21\textwidth}
		\centering
		\includegraphics[width=\textwidth]{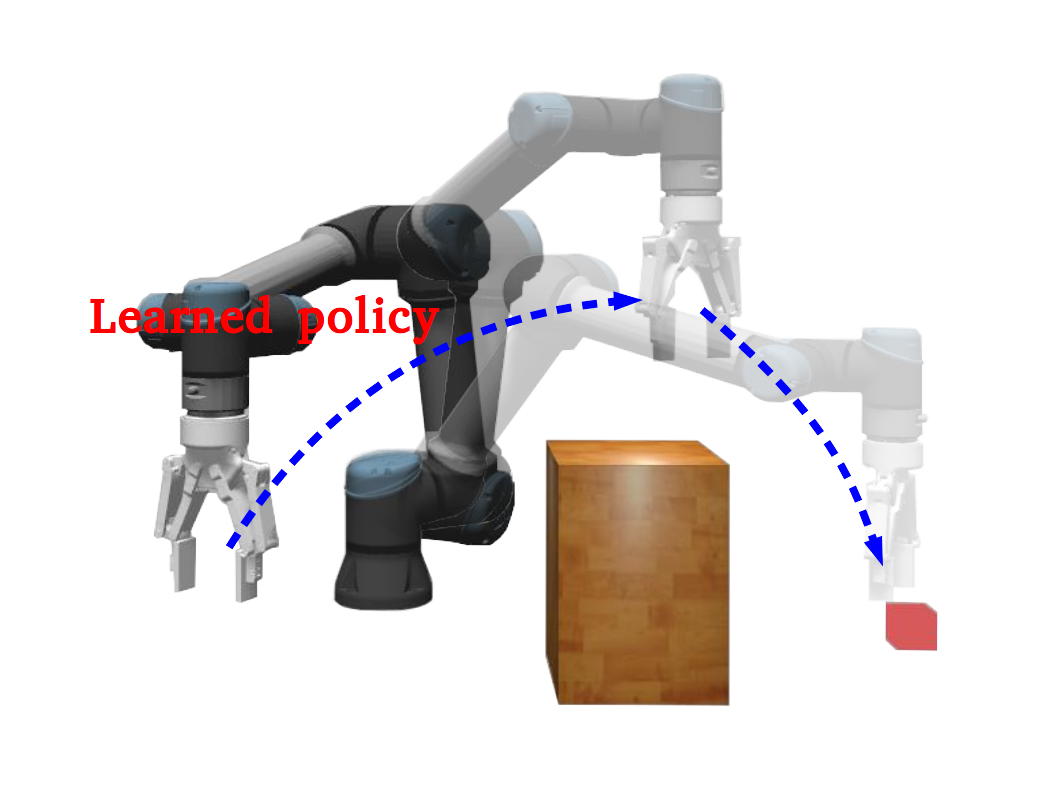}
		% \caption[Training in simulation]%
		% {{\small Training in simulation}}    
		\label{fig:place sim}
	\end{subfigure}
	%\quad
	%\hspace{.001in}
	\begin{subfigure}{0.21\textwidth}  
		\centering 
		\includegraphics[width=\textwidth]{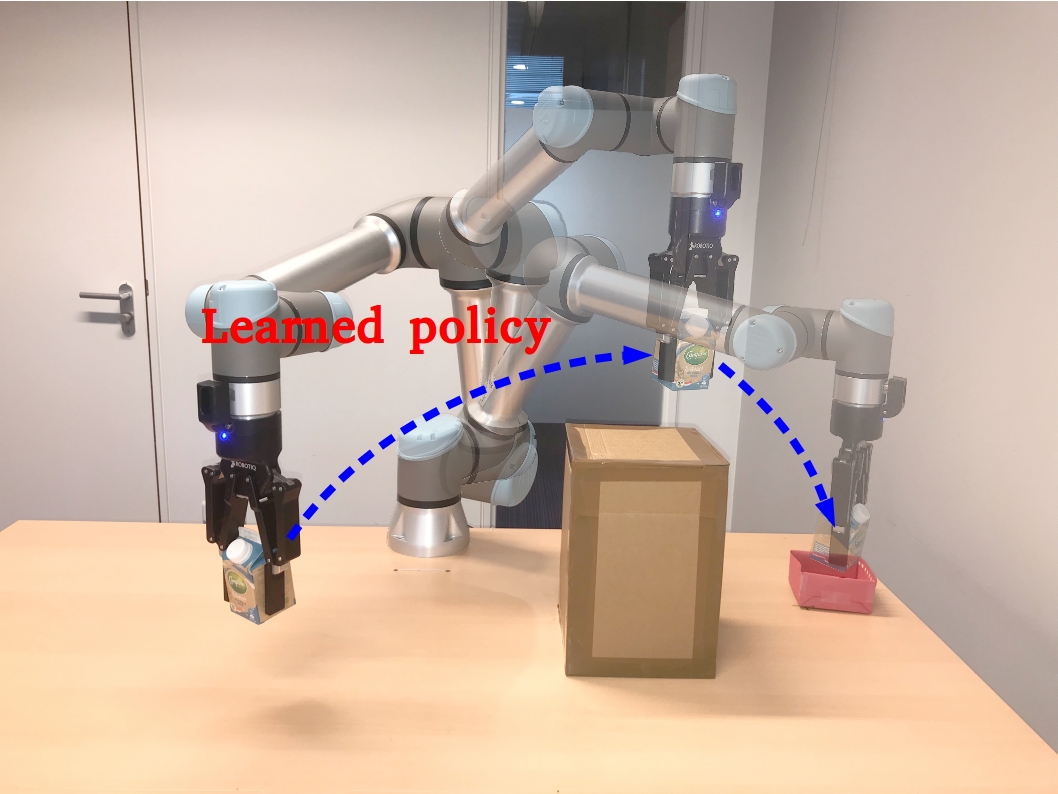}
		% \caption[Placement in real-world]%
		% {{\small Placement in real-world}}    
		\label{fig:place real}
	\end{subfigure}
	\vspace{-4mm}
	\caption[Placement task illustration]
	{\small Illustration of SILP trained in a simulator (left) and tested in a real-world placement task (right). With the learned policy, the robot places the object accurately while avoiding obstacles. } 
	\label{figplace}
	\vspace{-6mm}
\end{figure}

%%%%%%%%%%%%%%%%%%%%%%%%%%%%%%%%%%%%%%%%%%%%%%%%%%%%%%%%%%%%%%%%%%%%%%%%%%%%%%%%
\section{Related Work}

Following recent advances in deep learning, methods using neural networks to approximate motion planning algorithms have attracted increasing attention. Qureshi~et~al.~\cite{qureshi2019motion} learned a neural motion planner with supervised learning. Benjamin~et~al.~\cite{eysenbach2019search} and Xia~et~al.~\cite{xia2020relmogen} utilized hierarchical reinforcement learning to tackle long horizon motion planning tasks. Notably, Jurgenson~et~al.~\cite{DBLP:conf/rss/JurgensonT19} employed demonstrations with off-policy reinforcement learning to solve motion planning tasks and illustrated higher accuracy and less planning time compared with sampling-based motion planners. Their work is close to ours as we both utilize demonstrations in off-policy reinforcement learning. However, they focused on a simulated planar environment, but we consider a six degree-of-freedom (DoF) manipulator in a 3D task space. 

Learning from demonstrations (LfD) is a widely used technique for learning based robotic tasks~\cite{abbeel2007application, rigter2020framework, xiao2020appld, nair2017overcoming}, and behavior cloning (BC) is one of the most common frameworks in this paradigm, in which the robot learns by mimicking behaviors from the supervisor. However, in such a learning by imitation framework, the control error can accumulate with the time steps forward, leading the robot away from the familiar state region. The improved version of BC is DAgger~\cite{ross2011reduction}, which retrieves expert actions from the supervisor in all of the encountered states and aggregates the revised state-action pairs into the training data for later usage. The shortcoming of DAgger is the considerable burden on the supervisor who must label all the visited states during training in complex tasks. One common limitation of IL is that the agent cannot learn a policy better than the supervisor.

By contrast, reinforcement learning can discover better policies by exploring, but it suffers from slow convergence. Reinforcement learning with demonstrations can exploit the strength of RL and IL and overcome their respective weaknesses, leading to wide usage in complex robotics controlling tasks. In such frameworks, demonstrations mainly function as an initialization tool to boost RL policies, and expect the following exploration process to find a better policy than that of the supervisor~\cite{vecerik2017leveraging,hester2017deep,nair2017overcoming}. Aside from bootstrapping exploration in RL, demonstrations can be used to infer the reward function in reinforcement learning \cite{abbeel2004apprenticeship}, which belongs to the branch of inverse reinforcement learning~(IRL) and will not be covered in this study. 

Our approach is similar to the work introduced in~\cite{nair2017overcoming}, where demonstrations are stored in an extra replay buffer to guide the training by constructing an auxiliary behavior cloning loss to the policy update. We use the same framework but improve it with our SILP approach to reduce the required human effort in collecting demonstrations. Besides, our method is similar to DAgger as an online demonstration approach, which gives feedback for the encountered states by relabelling actions to minimize the behavior regret.  Compared to DAgger, our method has a global vision: it plans on all states within an episode rather than at one single step. In this manner, we can highlight the most important steps in the episode. Another work shares the concept of self-imitation learning \cite{oh2018self} with our method by utilizing the good decisions in the past episode to improve the learning performance. However, the quality of their method depends on the RL exploration strategies. When considering complex robotic tasks, it is difficult for the policy to obtain informative steps without explicit supervision. 

From the perspective of data augmentation, learning by imagination~\cite{racaniere2017imagination} aggregates data by rolling out the policy based on a learned environment model, which is an efficient way to accumulate useful data if an accurate model can be acquired. In addition, hindsight experience replay (HER) \cite{andrychowicz2017hindsight} aggregates informative experience by regarding the experienced next state as the goal. However, while more successful experiences are aggregated, the number of useless experiences grows as well and thus deteriorates the learning efficiency. Nevertheless, our method can select the most promising state as the next state and extract the core steps in the episode that lead to success. 

%%%%%%%%%%%%%%%%%%%%%%%%%%%%%%%%%%%%%%%%%%%%%%%%%%%%%%%%%%%%%%%%%%%%%%%%%%%%%%%%
\section{Preliminaries and Problem Formulation} \label{problemformulation}
\subsection{Preliminaries}
\textbf{Markov Decision Process:} A Markov Decision Process~(MDP) can be described as a tuple containing four basic elements: $(s_{t}, a_{t}, p(s_{t+1}|s_{t}, a_{t}), r(s_{t+1}|s_{t}, a_{t}))$. In this tuple, the $s_t$ and $a_t$ denote the continuous state and action at time step $t$ respectively, $p(s_{t+1}|s_{t}, a_{t})$ is the transition function of which the value is the probability to arrive at the next state $s_{t+1}$ given the current state $s_{t}$ and action $a_t$, and $r(s_{t+1}|s_{t}, a_{t})$ is the immediate reward received from the environment after the state transition. 

\textbf{Off-policy RL:} Different from offline RL where data is collected once in advance \cite{levine2020offline}, in online RL, an agent continuously updates data by interacting with the environment to learn the optimal policy $\pi^*$. The goal of the agent is to maximize the expected future return $R_{t}=\mathbb{E}[\sum_{i=t}^\infty\gamma^{i-t}r_{i+1}]$ with a discounted factor $\gamma \in [0, 1]$ weighting the future importance. Each policy $\pi$ has a corresponding action-value function $Q^{\pi}(s, a)=\mathbb{E}[R_t|s_t=s, a_t=a]$, representing the expected return under policy $\pi$ after taking action $a$ in state $s$. Following policy $\pi$, $Q^{\pi}$ can be computed through the Bellman equation:
\begin{equation}\label{Qfunc}
\begin{split}
\scalemath{0.87}{
Q^{\pi}(s_t, a_t)  = \mathbb{E}_{s_{t+1}\sim p}[r(s_t, a_t)+ \gamma\mathbb{E}_{a_{t+1}\sim{A}}[Q_{\pi}(s_{t+1}, a_{t+1})]],
}
\end{split}
\end{equation}
in which $A$ represents the action space. Let $Q^{*}(s, a)$ be the optimal action-value function. RL algorithms aim to find an optimal policy $\pi^*$ such that $Q^{\pi^*}(s, a)=Q^{*}(s, a)$ for all states and actions. 

In off-policy RL, the state-action pairs in \eqref{Qfunc} are not from the same policy. For instance, $a_t$ is from the current policy, while the next action $a_{t+1}$ may be a greedy action or comes from a previous policy. The latter case represents a line of work in off-policy RL that utilize an replay buffer to store interactions for future learning in order to remove the correlation between data \cite{lillicrap2015continuous, haarnoja2018soft1}. 

\subsection{Problem Formulation}
Given the collision-free start configuration $q_0$ and goal configuration $q_g$, the motion planning task is to find a path that starts from $q_0$ and ends with $q_g$, while avoiding obstacles $o$ in the environment. This task can be formulated as a fully observable Markov Decision Process~(MDP), which can be solved in the off-policy reinforcement learning framework. The detailed formulation is described below.

\subsubsection{\textbf{States}}
A feature vector is used to describe the continuous state, including the robot's proprioception, the obstacle and goal information in the environment. We restrict the orientation of the gripper as orthogonal and downward to the table, so three joints out of six in our UR5e platform are active in the learning process. At each step $t$ we record the $i$-th joint angles $j_i$ for $i=1,2,3$ in radians and the end-effector's position $(x^{ee}, y^{ee}, z^{ee}) \in \mathbb{R}^3$ as the proprioception: ${\rm proprio}=(j_1, j_2, j_3, x^{ee}, y^{ee}, z^{ee}) \in \mathbb{R}^6$. Then, we estimate the obstacle's position in task space and use a bounding box to describe it: ${\rm obs} = (x^{o}_{min}, x^{o}_{max}, y^{o}_{min}, y^{o}_{max}, z^{o}_{min}, z^{o}_{max}) \in \mathbb{R}^6$. The goal is described as a point in the task space: ${\rm goal}=(x^g, y^g, z^g) \in \mathbb{R}^3$. Finally, the state feature vector can be represented as: $s=({\rm proprio}, {\rm obs}, {\rm goal}) \in \mathbb{R}^{15}$. 
\subsubsection{\textbf{Actions}}
Each action is denoted by a vector $a \in ([-1, 1])^3$, which represents the relative position change for the first three joints. The corresponding three joint angle changes are $0.125 a$ rads.

\subsubsection{\textbf{Transition function}}
In each training episode, the goal and obstacle are static. So, the transition function is determined by the forward kinematics of the robot arm. Specifically, the next state~$s_{i+1}$ can be computed by the forward kinematics function~$f_s$ under the position controller; i.e., $s_{i+1} = f_s(s_i, a_i)$, where $s_{i}$ and $a_{i}$ are the current state and action respectively. Since the transition function is known, our off-policy reinforcement learning framework can also be seen as model-based. %Based on the kinematics model and position controller, the action between two states that in the same episode can be retrieved with the action model~$f_a$: $a_i = f_a(s_i, s_{i+1})$. However, the reward is unpredictable, since the collision information can only be accessed after the action executed. 

\subsubsection{\textbf{Rewards}}
A success is reached if the Euclidean distance between the end-effector and the goal ${\rm dis}({\rm ee}, g)<{\rm err}$, where ${\rm err}$ controls the reach accuracy. Given the current state and the taken action, if the next state is not collision-free, then a severe punishment is given by a negative reward $r=-10$. If the next state results in a success, we encourage such a behavior by setting the reward $r=1$. In other cases, $r=-{\rm dis}({\rm ee}, g)$ to penalize a long traveling distance. An episode is terminated when the predefined maximum steps or a success is reached. When a collision happens, we reset the agent and the environment one step back and randomly select a valid action to continue the episode.

We make a natural assumption that the goal states are reachable and collision-free. Note that although the transition function is deterministic, the reward is unpredictable since we do not implement the collision checking during training to reduce computation load. The collision information can only be obtained after the action has been executed. 
% the next state~$s_{i+1}$ is accessible, given the current state~$s_{i}$ and action~$a_{i}$, as it can be computed by the forward kinematics model~$f_s$: $s_{i+1} = f_s(s_i, a_i)$ under the position controller. Based on the kinematics model and position controller, the action between two states that in the same episode can be retrieved with the action model~$f_a$: $a_i = f_a(s_i, s_{i+1})$. However, the reward is unpredictable, since the collision information can only be accessed after the action executed. 

%%%%%%%%%%%%%%%%%%%%%%%%%%%%%%%%%%%%%%%%%%%%%%%%%%%%%%%%%%%%%%%%%%%%%%%%%%%%%%%%
\section{METHODS}

\subsection{RL with Demonstrations}
For learning from demonstrations with reinforcement learning, we adapt the approach from~\cite{nair2017overcoming}, which incorporates demonstrations into a separate replay buffer in off-policy RL. This results in two replay buffers, $D_{demo}$ and $D_{\pi}$, which stores demonstrations and real interactions respectively. In every update step, we draw $N_D$ examples from $D_{demo}$ and $N_{\pi}$ experiences from $D_{\pi}$ to train the policy.
To enable the demonstrations to guide policy learning, a behavior cloning loss is constructed as below:
\begin{equation}\label{loss bc}
L_{bc} = \sum\nolimits_{i=1}^{N_{D}} \,\norm{\pi(s_{i}|\theta^{\pi})-a_{i}}^{2},
\end{equation}
where $a_i$ and $s_i$ represent the action and state from $D_{demo}$ respectively, and $\theta^{\pi}$ represents the learning parameters in the policy. The constructed loss is added to the policy objective $J$ to push the policy to learn behaviors from the expert, weighted with hyperparameters $\lambda_1$ and $\lambda_2$. Furthermore, in order to avoid learning from imperfect demonstrations, a $Q_{filter}$ is employed to prevent adding behavior cloning loss when the policy's action is better than the action from the demonstrations. Therefore, the gradient applied to the policy parameter $\theta_\pi$ is:
\begin{equation}\label{loss policy}
\lambda_1\nabla_{\theta_{\pi}}J-\lambda_2 Q_{filter} \nabla_{\theta_\pi}L_{bc},
\end{equation}
where
\begin{equation}\label{Q filter}
Q_{filter} =\left\{
\begin{aligned}
& 1 & & \text{if}\quad Q(s_i, a_i)>Q(s_i, \pi(s_i)),\\
& 0 & & \text{otherwise}.
\end{aligned}
\right.
\end{equation}

\subsection{Self-imitation learning by planning}
Many graph-search based planning approaches can be used in our SILP to plan paths on the replay buffer; here we use probabilistic roadmap~(PRM)~\cite{kavraki1996probabilistic} as the planner. 
\subsubsection{Probabilistic Roadmap Construction} PRM is a multi-query sampling-based motion planning algorithm; it finds a path that connects the agent's start configuration and goal configuration while avoiding the obstacles. It contains two stages: a collision-free roadmap (graph) construction stage that randomly samples nodes in the configuration space and a path planning stage which plans on the constructed roadmap with a local planner.
We build a directed graph on the visited states $S$ in each episode. Therefore, each node corresponds to a state in the MDP. Then, we denote edges between nodes with states transition: $e_{s\rightarrow s'}$. Edges with Euclidean distance longer than $d$ are ignored. With this condition, we assume the state transition are safe in the graph. Finally, the graph $\mathcal{G}$ is formulated as follows:
\begin{equation}\label{roadmap}
\begin{split}
	 \mathcal{G} \triangleq (\mathcal{V}, \mathcal{E}) \quad \text{ where } \quad \mathcal{V} = \mathcal{S},\\
	 \mathcal{E} =\left\{
	\begin{aligned}
	& \emptyset & & d(s, s')>d , \\
	& \{e_{s\rightarrow s'}\mid s, s'\in \mathcal{S}\} & & \text{otherwise},
	\end{aligned}
	\right.
\end{split}
\end{equation}
where $\mathcal{V}$ and $\mathcal{E}$ represent the nodes and edges in the graph respectively.
Then, we use \textit{A-star} as the local planner to extract paths in each episode, in which the end effector's positions in states are used to calculate the heuristic function.

\subsubsection{Online Demonstrations Generation}
The pseudo-code for online demonstrations generation is shown in Alg.~\ref{PRM-algorithm}.
We denote $S_e=\{s_{e}^1, s_{e}^2, ...s_{e}^n\}$ as the visited states in episode~$e$, which can be seen as the collision-free nodes in PRM. Given the start node $s_0$ and the goal node $s_n$, we use \textit{A-star} to plan the shortest path between them. The extracted nodes that construct the demonstrated path is represented as: $V_e=\{s_0, ..., s_n\} \subseteq S_e$. After planning, the path is transferred into the MDP format for reinforcement learning. Under the position control and the known kinematics model, we can derive the action model function $f_a$ and use this function to compute the action from two neighboring states: $a_i = f_a(s_i, s_{i+1})$. If the action is out of the space $A$, we insert an extra node between the states with a half value of $a_i$, and the new next state is calculated with the forward kinematics function~$f_s$ defined in Sec.~\ref{problemformulation}. This process is repeated until the action locates within the defined space. We save the constructed demonstrations $(s_i, a_i, s_{i+1}, r_i)$ in $D_{demo}$ for policy learning. 

\begin{figure}[t]
	\centering
	\includegraphics[scale=0.66]{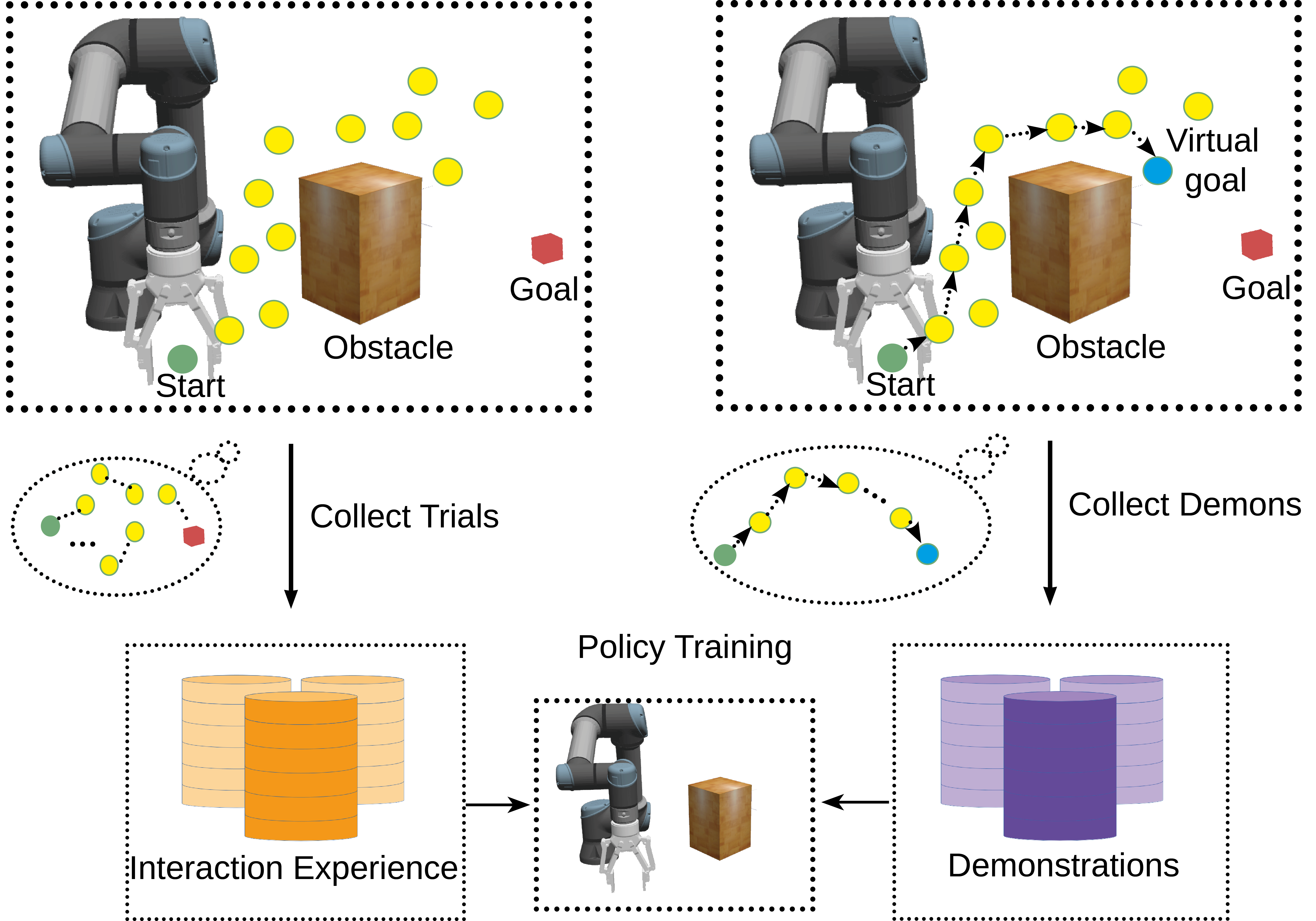}
	\caption[Self-imitation learning by planning]
	{\small Self-imitation learning by planning (SILP). %Top left: the visited states (yellow points) in an RL episode; Top right: the planned path (directed-dashed lines) that connects the start node (green point) and the virtual goal (blue point). 
	}
	\label{fig:PRM-architecture}
	\vspace{0mm}
\end{figure}
\setlength{\textfloatsep}{0pt}% Remove \textfloatsep
\begin{algorithm2e}[t]
	\textbf{Input}: Visited states $S_e = \{s_e^1, s_e^2, ..., s_e^n\}$ in episode~$e$; start and goal states $s_{0}$, $s_{n} \in S_e$; distance~$d$; action model function $f_a$ and forward kinematics function $f_s$\;
	\textbf{Output}: Updated demonstration replay buffer $D_{demo}$\;
	\textbf{Local planner}: Search the shortest path from $s_{0}$ to $s_{n}$: $V_e=\{s_0, ..., s_n\} \subseteq S_e$ \;
	\textbf{Demonstrations Generation}:\\
		\For{$i$ in \textsc{length}($V_e$)}
		{
			$s = V_e[i]$ \;
			$s' = V_e[i+1]$ \;
			$a = f_a(s, s')$ \;
			   \While{$a$ $\not\in$ $A$}{
			   	$(s'', a') = \textsc{InsertNodes}(s, s', a)$\;
				$r' = env.\textsc{reward}(s, s'')$ \;
				$D_{demo}.\textsc{push}((s, a', s'', r'))$\;
				$ s = s''$ \;
				$a = f_a(s'', s')$ \;
			   } 
			$r = env.\textsc{reward} (s, s')$ \;
			$D_{demo}.\textsc{push}((s, a, s', r))$
		}
		   \SetKwFunction{FMain}{InsertNodes}
		   \SetKwProg{Fn}{Function}{:}{}
		   \Fn{\FMain{$s$, $s'$, $a$}}{
		   	$a = a/2$\;
		   	$s' = f_s(s, a)$\;
		   	\While {$a$ $\not\in$ $A$}{
		   		$(s', a) = \textsc{InsertNodes}(s, s', a)$\\
		   	}
		   	\KwRet $s'$, $a$\;
		   }
	\caption{Demonstrations Generation in SILP}
	\label{PRM-algorithm}
	%\vspace{-2mm}
\end{algorithm2e}
The architecture of our SILP is shown in Fig.~\ref{fig:PRM-architecture}. From the top left sub-figure, we see that the original goal (red box) may be far from the explored states. This situation is common in the early stage of RL learning when the policy is not well trained. In this case, we cannot directly use PRM to plan a path from the start node to the goal node. Instead, we utilize HER~\cite{andrychowicz2017hindsight} to generate virtual goals from states that the agent has already visited (i.e., the blue point in the top right sub-figure in Fig.~\ref{fig:PRM-architecture}). An example of the planned path is depicted in Fig.~\ref{fig:PRM-architecture}~(top right), where the extracted nodes of the path are connected with the directed-dashed lines. 

Although we plan demonstrations concurrently with training, we do not add a substantial computation burden on the main algorithm. The reasons are: (1) We plan on the visited states, which eliminates the collision-checking process and accelerates the graph construction process; (2) The local planner occupies a small proportion of computation compared to the interaction with environment process in RL. 

%%%%%%%%%%%%%%%%%%%%%%%%%%%%%%%%%%%%%%%%%%%%%%%%%%%%%%%%%%%%%%%%%%%%%%%%%%%%%%%%
\section{Experiments and Results}
We conduct the experiments in both simulations and real-world settings to test our proposed method. The policies are learned in simulations due to safety consideration and then evaluated in both simulations and physical robot. We will investigate the following questions: (1) Under the same environment and task settings, does our method achieve better performance than other baselines in terms of success rate and sample efficiency? (2) Will our SILP method that has a online demonstration generation setting increases computation burden, leading to a slower training process? (3) Can the policy learned in simulation transfers well to a physical robot where noise and uncertainty exist?  
\subsection{Environment and tasks settings}
\begin{figure}[!t]
	%\vspace{-2mm}
	\centering
	\begin{subfigure}[t]{0.2\textwidth}
		\centering
		\includegraphics[width=\textwidth]{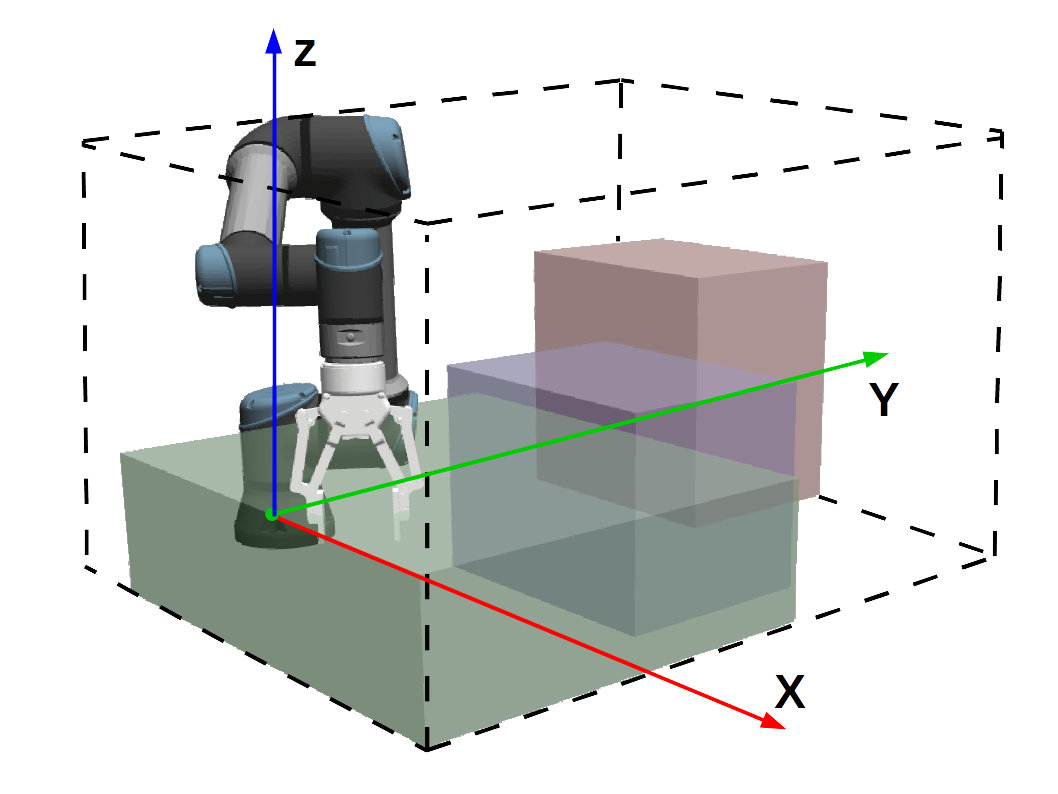}
		\caption[Workspace for Task 1]%
		{{\small Workspace for Task 1}}    
		\label{fig:task1}
	\end{subfigure}
	\quad
	\hspace{.01in}
	\begin{subfigure}[t]{0.18\textwidth}  
		\centering 
		\includegraphics[width=\textwidth]{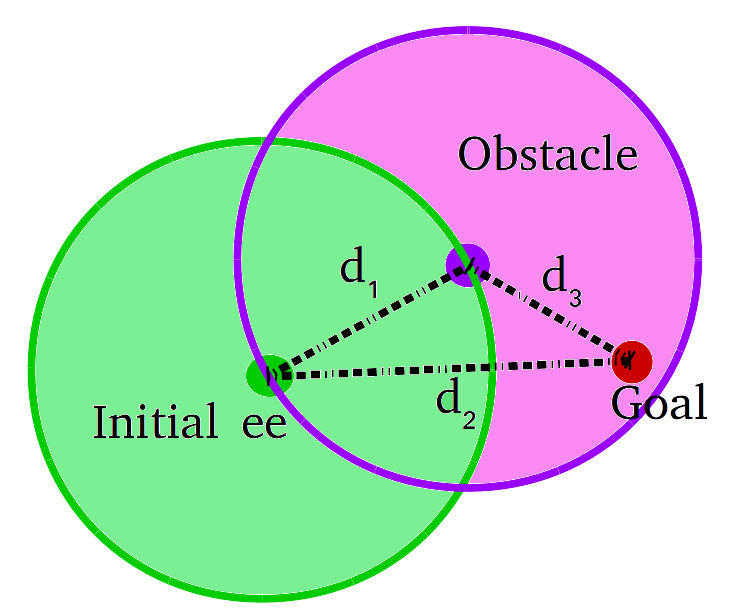}
		\caption[Workspace for Task 2]%
		{{\small Workspace for Task 2}}    
		\label{fig:task2}
	\end{subfigure}
	\caption[ ws illustration]
	{\small (a) Workspace for Task 1: the transparent region bounded with dashed line is the workspace of the end effector; the green box represents the initial end effector region; the purple region represents the obstacle's region; the goal is restricted in the red box. (b) Workspace for Task 2: the Euclidean distance between the goal, the obstacle and the initial end-effector's position satisfies \mbox{$d_2>d_1>d_3$}.} 
	\label{fig:ws}
	\vspace{0mm}
\end{figure}

We use Gazebo with an ODE physics engine to conduct our simulations, in which a 6 DoF robot arm UR5e is equipped with a Robotiq-2f-140 gripper to accomplish long horizon motion planning tasks. We design three tasks with ascending difficulty levels for the experiments. The workspace for the end-effector is restricted to $x\in[0, 0.8]m$, $y\in[-0.3, 0.8]m$, $z\in[0, 0.6]m$ to simplify the tasks and avoid unnecessary collision. A box with the width, height of 0.2m and 0.3m is used as an obstacle in these tasks. \\
\textbf{Task 1}: The initial pose of the arm is limited to the right side of the obstacle and the goal is limited to the other side of it to ensure enough collision experience during training. This means that there is one mode of behavior for the arm: moving from the right side of the obstacle to the left side while avoiding the obstacle. The workspace of this task is depicted in Fig.~\ref{fig:task1}, in which the initial end-effector's position (i.e., the center of the gripper's tip) is limited to $x\in[0.1, 0.8]m$, $y\in[-0.3, 0.3]m$, $z\in[0.0, 0.2]m$, the obstacle's position (i.e., the center) is limited to $x\in[0.3, 0.7]m$, $y\in[0.1, 0.4]m$, $z=0.15m$, and the goal's position is limited to $x\in[0.0, 0.4]m$, $y\in[0.5, 0.8]m$, $z\in[0, 0.4]m$. \\
\textbf{Task 2}: The initial and goal poses of the arm have the same space as the end effector. In order to balance the number of collision and non-collision interactions, we restrict the initial pose, goal pose and obstacle position to satisfy $d_2>d_1>d_3$, where $d_1$ is the Euclidean distance between the initial end-effector's position and the obstacle position, $d_2$ is the Euclidean distance between initial end-effector's position and the goal position and $d_3$ is the Euclidean distance between the obstacle position and the goal position. The robot needs to learn more and better to generalize to a bigger workspace that covers more modes. The relative position is projected in 2D and depicted in Fig.~\ref{fig:task2}. \\
\textbf{Task 3}: This task is designed based on \textbf{Task  2}, with two more obstacles in different shapes: $[0.3, 0.3]m$, $[0.3, 0.4]m$. Each shape contains the width and height of the obstacle. On each episode, one of the three boxes is randomly selected as the obstacle.
\subsection{Baselines and Implementation}
We employ two state-of-the-art off-policy RL algorithms as our basic baselines: deep deterministic policy gradient (DDPG) \cite{lillicrap2015continuous} and soft actor-critic (SAC) \cite{haarnoja2018soft1, haarnoja2018soft2} to evaluate our SILP method. We call them \textit{DDPG-SILP} and \textit{SAC-SILP} respectively. We consider the following algorithms as baselines to compare with:
\begin{itemize}
	\item \textbf{\textit{DDPG-baseline}}: DDPG combined with a dense reward.
	\item \textbf{\textit{SAC-baseline}}: SAC combined with a dense reward.
	\item \textbf{\textit{DDPG-HER}}: DDPG combined with hindsight experience replay (HER)~\cite{andrychowicz2017hindsight}. 
	\item \textbf{\textit{SAC-HER}}: SAC combined with HER.
	\item \textbf{\textit{BC}}: Behavior cloning with demonstrations collected from learned policies. 
\end{itemize}
The architectures of the neural networks and the hyperparameters of DDPG and SAC are illustrated in the \textit{Experiments} section in \cite{9207427}, except that we set the batch size as 256 and the total epochs as 1K in training. In SILP, we set the number of demonstrations $N_D$ as 32 in each batch training, then $N_{\pi}$ is 224. We set $\lambda_1=0.004$ and $\lambda_2=0.03125$ in \eqref{loss policy}. We choose to utilize the same configurations in \cite{9207427} as we have similar environments settings. For SAC, we initialize the policy by randomly exploring the environment with valid actions for the first $m$ epochs. We select $m=50$, which is the best one compared to 100, 150 and 200 exploration epochs. For the entropy coefficient, we compared three different configurations: the fixed value of 0.2, auto-tuning with dual variable optimization \cite{haarnoja2018soft2} and meta parameter learning \cite{wang2020meta}. We choose the auto-tuning method as it gained the best performance in terms of training time and success rate compared to the other two methods in a pilot experiment. 

For the baseline HER, the number of imagined goals is four for each visited states. We use 10$k$ demonstrations to train the BC model and the success rate in BC is an average of five randomly selected seeds. The demonstrations in BC are collected from the learned policies from the baselines mentioned before, as well as our methods DDPG-SILP and SAC-SILP. Each method contributes the same number of demonstrations to train the BC model. 
\subsection{Simulation experiments}
\begin{figure}[t!]
	%\vspace{-5mm}
	\centering
	%\vspace{-2mm}
	\begin{subfigure}[t]{0.5\textwidth}
		\centering
		\includegraphics[width=\textwidth]{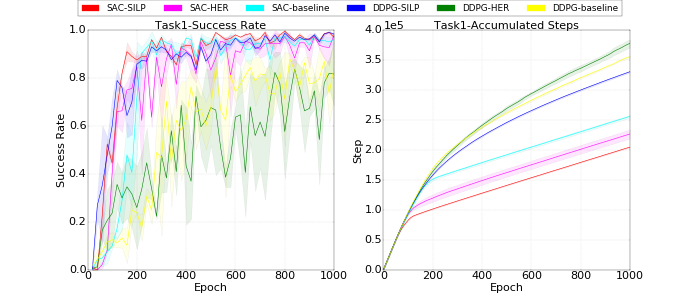}
	%	\caption[Task1]%
		%{{\small Task1}}    
	%	\label{fig:success1}
	\end{subfigure}
	%\quad
	%\hspace{.01in}
	%\vspace{-2mm}
	\begin{subfigure}[t]{0.5\textwidth}  
		\centering 
		\includegraphics[width=\textwidth]{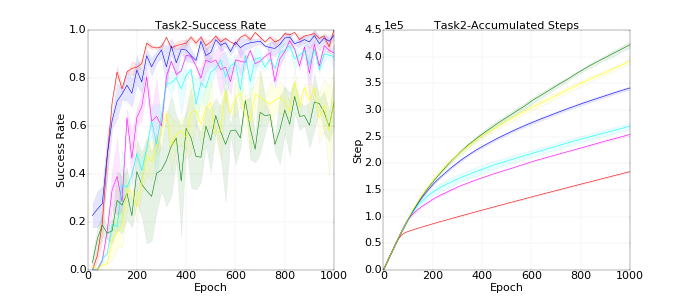}
		%\caption[Task2]%
		%{{\small Task2}}    
	%	\label{fig:success2}
	\end{subfigure}
	%\quad
	%\hspace{.01in}
	\begin{subfigure}[t]{0.5\textwidth}  
		\centering 
		\includegraphics[width=\textwidth]{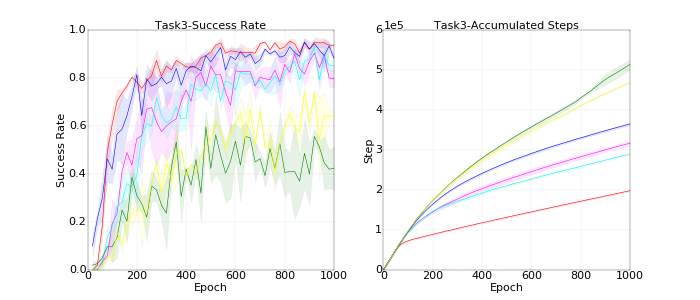}
		%\caption[Task3]%
		%{{\small Task3}}    
	%	\label{fig:success3}
	\end{subfigure}	
	\caption[question1]
	{\small Success rate (left column) and accumulated steps (right column) during the training for Task 1 (first row), Task 2 (second row) and Task 3 (third row); curves represent the mean value over five randomly-seeded runs, bounded with the variance represented by semi-transparent colors. Data are recorded every 20 epochs.} 
	\label{fig:success rate}
	\vspace{-3mm}
\end{figure}

To evaluate the sample efficiency and performance, we draw the curves of the success rate and accumulated steps during training in Fig.~\ref{fig:success rate}. From the figures on the right column, we can see that our methods, DDPG-SILP (blue curve) and SAC-SILP (red curve), use the least steps in all of the three tasks compared to other DDPG-based and SAC-based methods. Especially in Task 2 and Task 3, our SILP methods reduce the total steps by around 34\%. The figures on the left column demonstrate that SAC-based algorithms perform better than DDPG-based methods, but our methods can boost DDPGs' performance to reach the same level as SACs'. Especially for Task 3, DDPG-SILP improves the success rate to around 90\% compared to the success rate of around 65\% and 50\% for DDPG-baseline and DDPG-HER respectively, which is also illustrated in Table. \ref{tab:success_rate}. 

To compare the final performance of the learned policies, we tested the final success rate of SILP and other baselines, and summarized the average results in Table. \ref{tab:success_rate}. The results demonstrate that our methods achieve the highest final success rate in all of the three tasks. The difference between our approach and others is larger when the task is more difficult.

\begin{table}[!t]
	\vspace{-2mm}
	%\tiny
	\centering
	\caption {Training time and final mean success rate (SR).}
	\begin{tabular}{|c|c|c|c|c|c|c|c|}%
		\hline
		\multirow{2}*{Methods}&\multicolumn{2}{|c|}{Task1}&\multicolumn{2}{|c|}{Task2}&\multicolumn{2}{|c|}{Task3}\\
		\cline{2-7}
		~&SR(\%)&Time(h)&SR &Time&SR &Time\\
		\hline
		DDPG-baseline&84.2&11.86&79.0&14.14&68.0&16.9\\
		DDPG-HER&88.6&13&79.4&14.95&50.0&17.2\\
		\textbf{DDPG-SILP}&\textbf{98.8}&\textbf{11.15}&\textbf{97.4}&\textbf{13.12}&\textbf{92.0}&\textbf{14.9}\\
		SAC-baseline&97.6&9.16&86.4&10.61&89.0&11.54\\
		SAC-HER&96.6&9.65&87.2&10.26&83.6&11.53\\
		\textbf{SAC-SILP}&\textbf{98.6}&\textbf{7.38}&\textbf{97.6}&\textbf{8.43}&\textbf{93.8}&\textbf{9.33}\\
		BC&87.5&*&72.5&*&70.07&*\\
		\hline
	\end{tabular}
    \label{tab:success_rate}%
    \vspace{0mm}
\end{table} 
\begin{figure}[tb]
%	\vspace{-4mm}
	\centering
		\begin{subfigure}[t]{0.15\textwidth}  
		\centering 
		\includegraphics[width=\textwidth]{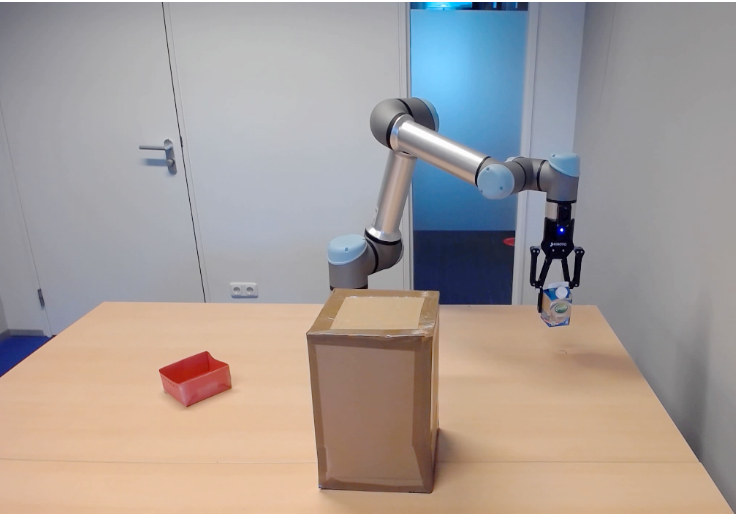}
	\end{subfigure}
	\begin{subfigure}[t]{0.15\textwidth}  
		\centering 
		\includegraphics[width=\textwidth]{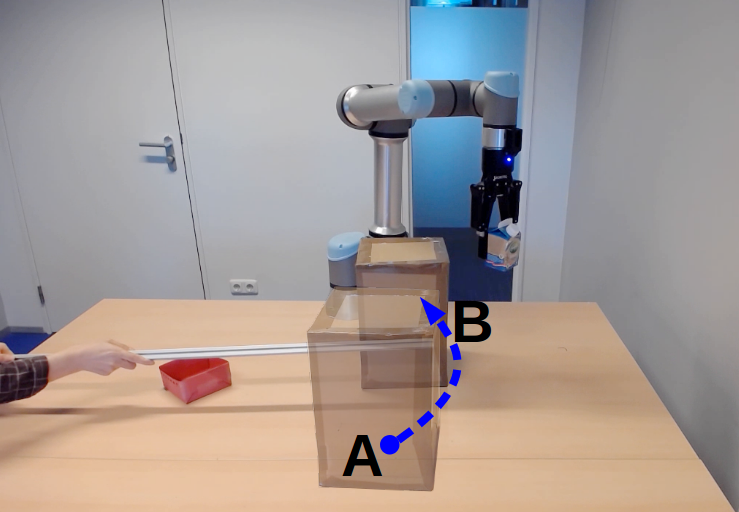}
	\end{subfigure}
	%\quad
	%\hspace{.01in}
	\begin{subfigure}[t]{0.15\textwidth}  
		\centering 
		\includegraphics[width=\textwidth]{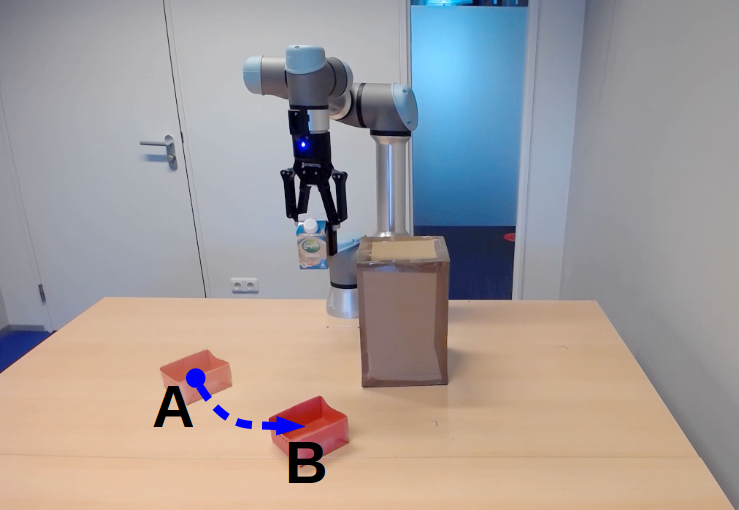}
	\end{subfigure}
	%\quad
	%\hspace{.01in}
	\caption[dynamic exp]
	{\small Placement task in a dynamic environment. (Left) Initial state; (Center) The obstacle is moved from point A to point B, interfering with the robot movement; (Right) The goal is moved from point A to point B.} 
	\label{fig:dynamic exp}
	\vspace{0mm}
\end{figure}
\begin{figure}
    \centering
    \includegraphics[width=0.3\textwidth]{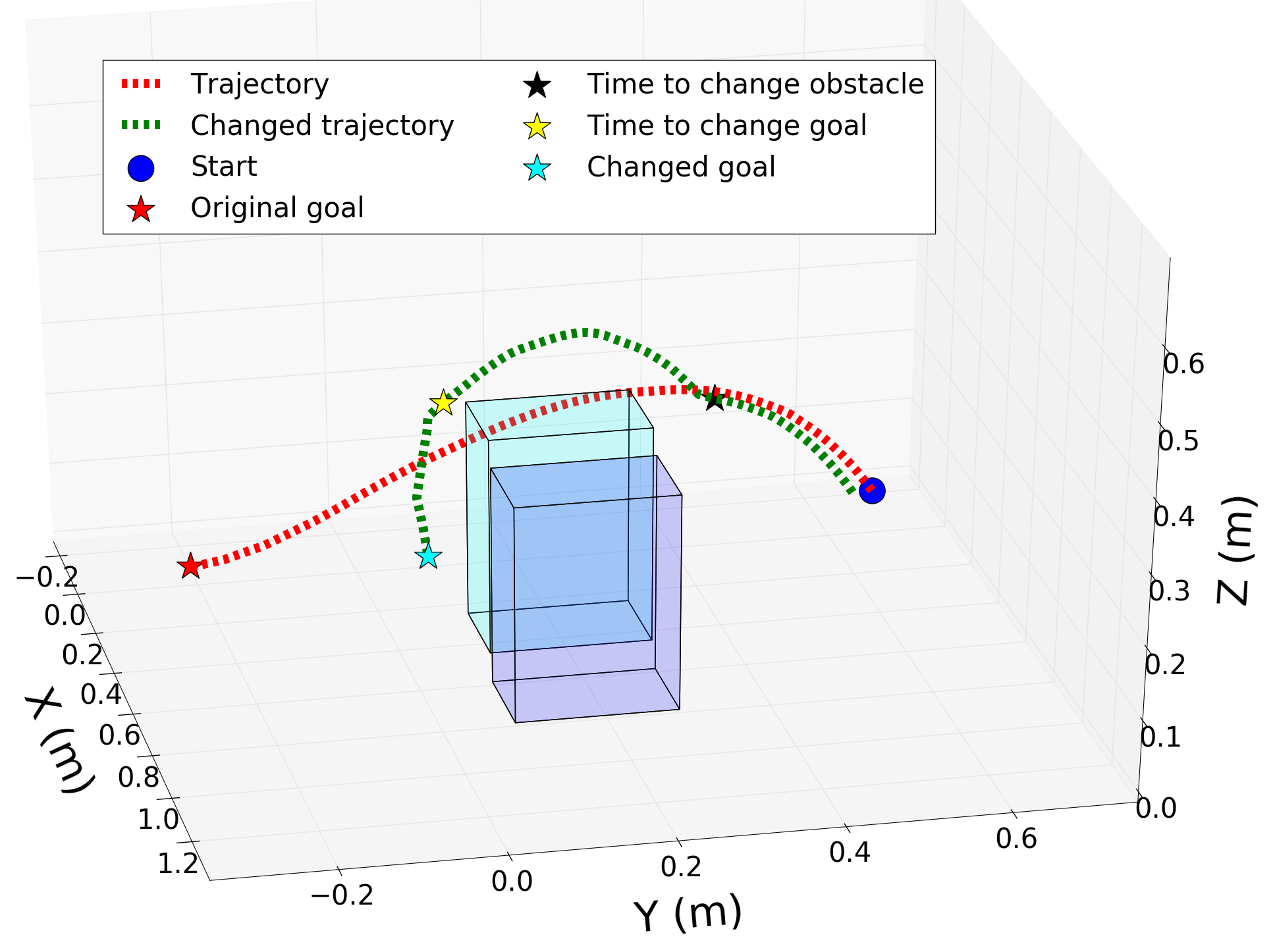}
    \caption[trajectory]
	{\small Trajectories of the end-effector in a dynamic environment. The purple and blue boxes represent the obstacle's positions before and after movement.} 
	\label{fig:trajectories}
	\vspace{0mm}
\end{figure}
To test the computation burden of our online demonstrations generation method, we summarized the total training time in Table. \ref{tab:success_rate}. From the table, we can see that our methods DDPG-SILP and SAC-SILP need around 13\% and 21\% less time compared to other DDPG-based and SAC-based methods for training the same number of epochs, even under the online planning setting. 

We also tested the time needed for pure PRM planning and compared it to the time for action retrieving in an episode with SILP-SAC under task2. We set the sampling nodes to 100 in PRM to realize an average success rate of 80\%. The average time for PRM and SILP-SAC are 18.988s and 0.131s respectively under five randomly-seeded runs on Intel-i7-6700 CPU processor running at 3.40GHz. From the result we can see that PRM is prohibitively computational heavy for multi-DoFs robot arm. The neural-based motion planner gains an advantage in terms of the computation load. 

In addition, the assumption that the movement between two neighboring states is collision-free in the planning phase is not always satisfied given noise and uncertainty. The recorded total collision numbers are higher in our methods compared to the baselines. However, Fig. \ref{fig:success rate} and Table. \ref{tab:success_rate} show that our method overcomes the imperfect assumption by the algorithm and improves the policy continuously, resulting in the highest success rate, the least-needed update steps and training time in tested tasks.
\subsection{Real-world experiments}
From the previous simulation results, we verified that our SAC-SILP method performs better than other methods in different tasks. In this section, we aim to evaluate the \textit{sim-to-real} ability of SAC-SILP on Task 2. We test the policy learned in simulation on a real-world placement application, which is shown in Fig.~\ref{figplace} and Fig.~\ref{fig:dynamic exp}. The obstacle is represented by a cardboard box that should be avoided during the movement. The object that needs to be grasped and placed is a milk bottle and the goal is a red basket to contain the object. These objects are randomly put on the table, detected and localized by the RACE real time system~\cite{kasaei2018towards, kasaei2019interactive}. The initial pose of the arm is randomly set up near the object. Once RACE detects and locates these objects, the robot moves to the milk bottle and grasps it, then the learned policy guides the arm towards the goal while avoiding the obstacle, and drop the object when a success is reached\footnote{see \url{https://youtu.be/2rx-roLYJ5k}}.
\subsubsection{Static environment}
We created 15 different configurations of the placement task by randomly scattering the obstacle, milk bottle and the red basket on the table to see how the learned policy places the milk bottle into the target basket while not colliding with the obstacle. We got 14 successes among 15 trials, which is reasonable when we consider the pose-estimation errors for these objects and the model difference between the simulated and the real robot.    
\subsubsection{Dynamic environment}
We tested the robustness of our policy in a dynamic setting, where the obstacle and goal were moving during the policy execution. The experiment is illustrated in Fig.~\ref{fig:dynamic exp}. In the original configuration, the obstacle is not blocking the path, which is easy for the robot arm to reach the goal, as shown in the left figure in Fig.~\ref{fig:dynamic exp}. During the execution, we move the obstacle to block the way to test the collision avoidance behavior as illustrated in the middle of Fig.~\ref{fig:dynamic exp}. Then, we change the goal position to test the adaptability of the learned policy as shown on the right of Fig.~\ref{fig:dynamic exp}. The original and changed trajectories are depicted in Fig.~\ref{fig:trajectories} by the dotted red and green curves, from which we can see obvious motion adaption when the environment changes. The observation shows that the learned policy reacts well in a dynamic environment, even it was trained in simulation. 
%%%%%%%%%%%%%%%%%%%%%%%%%%%%%%%%%%%%%%%%%%%%%%%%%%%%%%%%%%%%%%%%%%%%%%%%%%%%%%%%
\section{Conclusion and Future Work}

We present our self-imitation learning by planning (SILP) method, which relieves human effort on collecting diverse demonstrations in goal-conditioned long horizon motion planning tasks. With the guidance of self-generated demonstrations, we learn a policy that generalizes substantially better than those learned directly from RL, behavior cloning and HER, while not adding much extra computation burden on the training process. 

Although we tested our method with a position controller on a UR5e robot arm, SILP can also be used with other controllers and on other robotic platforms if an action model can be obtained to extract the MDP format demonstrations based on the explored states. In addition, we plan demonstrations in every episode, which adds unnecessary demonstrations when the policy performs well. This can be improved with a module to evaluate the policy uncertainty on visited states and decide if demonstrations are needed.
\clearpage
\addtolength{\textheight}{-12cm}   % This command serves to balance the column lengths
                                  % on the last page of the document manually. It shortens
                                  % the textheight of the last page by a suitable amount.
                                  % This command does not take effect until the next page
                                  % so it should come on the page before the last. Make
                                  % sure that you do not shorten the textheight too much.

%%%%%%%%%%%%%%%%%%%%%%%%%%%%%%%%%%%%%%%%%%%%%%%%%%%%%%%%%%%%%%%%%%%%%%%%%%%%%%%%

\bibliographystyle{IEEEtran}
\small{\bibliography{ref}}

\begin{thebibliography}{10}
\providecommand{\url}[1]{#1}
\csname url@rmstyle\endcsname
\providecommand{\newblock}{\relax}
\providecommand{\bibinfo}[2]{#2}
\providecommand\BIBentrySTDinterwordspacing{\spaceskip=0pt\relax}
\providecommand\BIBentryALTinterwordstretchfactor{4}
\providecommand\BIBentryALTinterwordspacing{\spaceskip=\fontdimen2\font plus
\BIBentryALTinterwordstretchfactor\fontdimen3\font minus
  \fontdimen4\font\relax}
\providecommand\BIBforeignlanguage[2]{{%
\expandafter\ifx\csname l@#1\endcsname\relax
\typeout{** WARNING: IEEEtran.bst: No hyphenation pattern has been}%
\typeout{** loaded for the language `#1'. Using the pattern for}%
\typeout{** the default language instead.}%
\else
\language=\csname l@#1\endcsname
\fi
#2}}

\bibitem{abbeel2008apprenticeship}
P.~Abbeel, D.~Dolgov, A.~Y. Ng, and S.~Thrun, ``Apprenticeship learning for
  motion planning with application to parking lot navigation,'' in \emph{2008
  IEEE/RSJ International Conference on Intelligent Robots and Systems}.\hskip
  1em plus 0.5em minus 0.4em\relax IEEE, 2008, pp. 1083--1090.

\bibitem{chiang2019learning}
H.-T.~L. Chiang, A.~Faust, M.~Fiser, and A.~Francis, ``Learning navigation
  behaviors end-to-end with autorl,'' \emph{IEEE Robotics and Automation
  Letters}, vol.~4, no.~2, pp. 2007--2014, 2019.

\bibitem{rosenbaum2001posture}
D.~A. Rosenbaum, R.~J. Meulenbroek, J.~Vaughan, and C.~Jansen, ``Posture-based
  motion planning: applications to grasping.'' \emph{Psychological review},
  vol. 108, no.~4, p. 709, 2001.

\bibitem{8794317}
C.~{Chamzas}, A.~{Shrivastava}, and L.~E. {Kavraki}, ``Using local experiences
  for global motion planning,'' in \emph{2019 International Conference on
  Robotics and Automation (ICRA)}, 2019, pp. 8606--8612.

\bibitem{4543471}
{Ruijie He}, S.~{Prentice}, and N.~{Roy}, ``Planning in information space for a
  quadrotor helicopter in a gps-denied environment,'' in \emph{2008 IEEE
  International Conference on Robotics and Automation}, 2008, pp. 1814--1820.

\bibitem{DBLP:conf/rss/JurgensonT19}
\BIBentryALTinterwordspacing
T.~Jurgenson and A.~Tamar, ``Harnessing reinforcement learning for neural
  motion planning,'' in \emph{Robotics: Science and Systems XV, University of
  Freiburg, Freiburg im Breisgau, Germany, June 22-26, 2019}, A.~Bicchi,
  H.~Kress{-}Gazit, and S.~Hutchinson, Eds., 2019. [Online]. Available:
  \url{https://doi.org/10.15607/RSS.2019.XV.026}
\BIBentrySTDinterwordspacing

\bibitem{qureshi2019motion}
A.~H. Qureshi, A.~Simeonov, M.~J. Bency, and M.~C. Yip, ``Motion planning
  networks,'' in \emph{2019 International Conference on Robotics and Automation
  (ICRA)}.\hskip 1em plus 0.5em minus 0.4em\relax IEEE, 2019, pp. 2118--2124.

\bibitem{ravichandar2020recent}
H.~Ravichandar, A.~S. Polydoros, S.~Chernova, and A.~Billard, ``Recent advances
  in robot learning from demonstration,'' \emph{Annual Review of Control,
  Robotics, and Autonomous Systems}, vol.~3, 2020.

\bibitem{ross2010efficient}
S.~Ross and D.~Bagnell, ``Efficient reductions for imitation learning,'' in
  \emph{Proceedings of the thirteenth international conference on artificial
  intelligence and statistics}, 2010, pp. 661--668.

\bibitem{chitta2012moveit}
S.~Chitta, I.~Sucan, and S.~Cousins, ``Moveit![ros topics],'' \emph{IEEE
  Robotics \& Automation Magazine}, vol.~19, no.~1, pp. 18--19, 2012.

\bibitem{rajeswaran2017learning}
A.~Rajeswaran, V.~Kumar, A.~Gupta, G.~Vezzani, J.~Schulman, E.~Todorov, and
  S.~Levine, ``Learning complex dexterous manipulation with deep reinforcement
  learning and demonstrations,'' \emph{arXiv preprint arXiv:1709.10087}, 2017.

\bibitem{sauser2012iterative}
E.~L. Sauser, B.~D. Argall, G.~Metta, and A.~G. Billard, ``Iterative learning
  of grasp adaptation through human corrections,'' \emph{Robotics and
  Autonomous Systems}, vol.~60, no.~1, pp. 55--71, 2012.

\bibitem{kober2009policy}
J.~Kober and J.~R. Peters, ``Policy search for motor primitives in robotics,''
  in \emph{Advances in neural information processing systems}, 2009, pp.
  849--856.

\bibitem{kalashnikov2018qt}
D.~Kalashnikov, A.~Irpan, P.~Pastor, J.~Ibarz, A.~Herzog, E.~Jang, D.~Quillen,
  E.~Holly, M.~Kalakrishnan, V.~Vanhoucke, \emph{et~al.}, ``Qt-opt: Scalable
  deep reinforcement learning for vision-based robotic manipulation,''
  \emph{arXiv preprint arXiv:1806.10293}, 2018.

\bibitem{eysenbach2019search}
B.~Eysenbach, R.~R. Salakhutdinov, and S.~Levine, ``Search on the replay
  buffer: Bridging planning and reinforcement learning,'' in \emph{Advances in
  Neural Information Processing Systems}, 2019, pp. 15\,246--15\,257.

\bibitem{xia2020relmogen}
F.~Xia, C.~Li, R.~Mart{\'\i}n-Mart{\'\i}n, O.~Litany, A.~Toshev, and
  S.~Savarese, ``Relmogen: Leveraging motion generation in reinforcement
  learning for mobile manipulation,'' \emph{arXiv preprint arXiv:2008.07792},
  2020.

\bibitem{abbeel2007application}
P.~Abbeel, A.~Coates, M.~Quigley, and A.~Y. Ng, ``An application of
  reinforcement learning to aerobatic helicopter flight,'' in \emph{Advances in
  neural information processing systems}, 2007, pp. 1--8.

\bibitem{rigter2020framework}
M.~Rigter, B.~Lacerda, and N.~Hawes, ``A framework for learning from
  demonstration with minimal human effort,'' \emph{IEEE Robotics and Automation
  Letters}, vol.~5, no.~2, pp. 2023--2030, 2020.

\bibitem{xiao2020appld}
X.~Xiao, B.~Liu, G.~Warnell, J.~Fink, and P.~Stone, ``Appld: Adaptive planner
  parameter learning from demonstration,'' \emph{arXiv preprint
  arXiv:2004.00116}, 2020.

\bibitem{nair2017overcoming}
A.~Nair, B.~McGrew, M.~Andrychowicz, W.~Zaremba, and P.~Abbeel, ``Overcoming
  exploration in reinforcement learning with demonstrations. corr
  abs/1709.10089 (2017),'' \emph{arXiv preprint arXiv:1709.10089}, 2017.

\bibitem{ross2011reduction}
S.~Ross, G.~Gordon, and D.~Bagnell, ``A reduction of imitation learning and
  structured prediction to no-regret online learning,'' in \emph{Proceedings of
  the fourteenth international conference on artificial intelligence and
  statistics}, 2011, pp. 627--635.

\bibitem{vecerik2017leveraging}
M.~Vecerik, T.~Hester, J.~Scholz, F.~Wang, O.~Pietquin, B.~Piot, N.~Heess,
  T.~Roth{\"o}rl, T.~Lampe, and M.~Riedmiller, ``Leveraging demonstrations for
  deep reinforcement learning on robotics problems with sparse rewards,''
  \emph{arXiv preprint arXiv:1707.08817}, 2017.

\bibitem{hester2017deep}
T.~Hester, M.~Vecerik, O.~Pietquin, M.~Lanctot, T.~Schaul, B.~Piot, D.~Horgan,
  J.~Quan, A.~Sendonaris, G.~Dulac-Arnold, \emph{et~al.}, ``Deep q-learning
  from demonstrations,'' \emph{arXiv preprint arXiv:1704.03732}, 2017.

\bibitem{abbeel2004apprenticeship}
P.~Abbeel and A.~Y. Ng, ``Apprenticeship learning via inverse reinforcement
  learning,'' in \emph{Proceedings of the twenty-first international conference
  on Machine learning}, 2004, p.~1.

\bibitem{oh2018self}
J.~Oh, Y.~Guo, S.~Singh, and H.~Lee, ``Self-imitation learning,'' in
  \emph{International Conference on Machine Learning}.\hskip 1em plus 0.5em
  minus 0.4em\relax PMLR, 2018, pp. 3878--3887.

\bibitem{racaniere2017imagination}
S.~Racani{\`e}re, T.~Weber, D.~Reichert, L.~Buesing, A.~Guez, D.~J. Rezende,
  A.~P. Badia, O.~Vinyals, N.~Heess, Y.~Li, \emph{et~al.},
  ``Imagination-augmented agents for deep reinforcement learning,'' in
  \emph{Advances in neural information processing systems}, 2017, pp.
  5690--5701.

\bibitem{andrychowicz2017hindsight}
M.~Andrychowicz, F.~Wolski, A.~Ray, J.~Schneider, R.~Fong, P.~Welinder,
  B.~McGrew, J.~Tobin, O.~P. Abbeel, and W.~Zaremba, ``Hindsight experience
  replay,'' in \emph{Advances in neural information processing systems}, 2017,
  pp. 5048--5058.

\bibitem{levine2020offline}
S.~Levine, A.~Kumar, G.~Tucker, and J.~Fu, ``Offline reinforcement learning:
  Tutorial, review, and perspectives on open problems,'' \emph{arXiv preprint
  arXiv:2005.01643}, 2020.

\bibitem{lillicrap2015continuous}
T.~P. Lillicrap, J.~J. Hunt, A.~Pritzel, N.~Heess, T.~Erez, Y.~Tassa,
  D.~Silver, and D.~Wierstra, ``Continuous control with deep reinforcement
  learning,'' \emph{arXiv preprint arXiv:1509.02971}, 2015.

\bibitem{haarnoja2018soft1}
T.~Haarnoja, A.~Zhou, P.~Abbeel, and S.~Levine, ``Soft actor-critic: Off-policy
  maximum entropy deep reinforcement learning with a stochastic actor,''
  \emph{arXiv preprint arXiv:1801.01290}, 2018.

\bibitem{kavraki1996probabilistic}
L.~E. Kavraki, P.~Svestka, J.-C. Latombe, and M.~H. Overmars, ``Probabilistic
  roadmaps for path planning in high-dimensional configuration spaces,''
  \emph{IEEE transactions on Robotics and Automation}, vol.~12, no.~4, pp.
  566--580, 1996.

\bibitem{haarnoja2018soft2}
T.~Haarnoja, A.~Zhou, K.~Hartikainen, G.~Tucker, S.~Ha, J.~Tan, V.~Kumar,
  H.~Zhu, A.~Gupta, P.~Abbeel, \emph{et~al.}, ``Soft actor-critic algorithms
  and applications,'' \emph{arXiv preprint arXiv:1812.05905}, 2018.

\bibitem{9207427}
S.~{Luo}, H.~{Kasaei}, and L.~{Schomaker}, ``Accelerating reinforcement
  learning for reaching using continuous curriculum learning,'' in \emph{2020
  International Joint Conference on Neural Networks (IJCNN)}, July 2020, pp.
  1--8.

\bibitem{wang2020meta}
Y.~Wang and T.~Ni, ``Meta-sac: Auto-tune the entropy temperature of soft
  actor-critic via metagradient,'' \emph{arXiv preprint arXiv:2007.01932},
  2020.

\bibitem{kasaei2018towards}
S.~H. Kasaei, M.~Oliveira, G.~H. Lim, L.~S. Lopes, and A.~M. Tom{\'e},
  ``Towards lifelong assistive robotics: A tight coupling between object
  perception and manipulation,'' \emph{Neurocomputing}, vol. 291, pp. 151--166,
  2018.

\bibitem{kasaei2019interactive}
S.~H. Kasaei, N.~Shafii, L.~S. Lopes, and A.~M. Tom{\'e}, ``Interactive
  open-ended object, affordance and grasp learning for robotic manipulation,''
  in \emph{2019 IEEE/RSJ International Conference on Robotics and Automation
  (ICRA)}.\hskip 1em plus 0.5em minus 0.4em\relax IEEE, 2019, pp. 3747--3753.

\end{thebibliography}
\end{document}